\documentclass[10pt,letterpaper]{article}

\usepackage{cogsci}

\cogscifinalcopy % Uncomment this line for the final submission 

\usepackage[
    backend=biber,
    style=apa,
    natbib=true,
    doi=false,
    isbn=false,
    url=false,
]{biblatex}
\usepackage{pslatex}
\usepackage{nicefrac}
\usepackage{float} % Roger Levy added this and changed figure/table
\usepackage{multirow}
\usepackage{booktabs}
\usepackage{tabularray}
\usepackage{graphicx}
\usepackage{subcaption}
\usepackage{amsmath,amssymb,amsthm,amsfonts}
\graphicspath{ {./Figure/} }
\usepackage{color}

\newcommand{\key}{\textbf}

\title{An information-theoretic model of shallow and deep language comprehension}
 
\author{{\large \bf Jiaxuan Li (jiaxuan.li@uci.edu)} \\
  Department of Language Science, University of California Irvine\\ Irvine, CA 92617 USA 
  \AND {\large \bf Richard Futrell (rfutrell@uci.edu)} \\
  Department of Language Science, University of California Irvine\\ Irvine, CA 92617 USA }

\begin{document}

\maketitle

\begin{abstract}

A large body of work in psycholinguistics has focused on the idea that online language comprehension can be shallow or `good enough': given constraints on time or available computation, comprehenders may form interpretations of their input that are plausible but inaccurate. However, this idea has not yet been linked with formal theories of computation under resource constraints. Here we use information theory to formulate a model of language comprehension as an optimal trade-off between accuracy and processing depth, formalized as bits of information extracted from the input, which increases with processing time. The model provides a measure of processing effort as the change in processing depth, which we link to EEG signals and reading times. We validate our theory against a large-scale dataset of garden path sentence reading times, and EEG experiments featuring N400, P600 and biphasic ERP effects. By quantifying the timecourse of language processing as it proceeds from shallow to deep, our model provides a unified framework to explain behavioral and neural signatures of language comprehension.

\end{abstract}

\section{Introduction}
Language comprehension can be inaccurate in the sense that a comprehender's interpretations of their input may not reflect the input veridically. For example, given input such as
\begin{itemize}
    \item[(1)] The storyteller could turn any incident into an amusing \emph{antidote} \dots
\end{itemize}
a comprehender may initially interpret the anomalous word \emph{antidote} as the more plausible \emph{anecdote}, and the misinterpretation may persist \citep{fillenbaum1971processing,fillenbaum1974pragmatic,erickson1981words,barton_case_1993,sanford2002context,ferreira2002good,ferreira_misinterpretation_2003}. Similarly, processing of syntactic structure can be inaccurate, with comprehenders interpreting a sentence such as 
\begin{itemize}
    \item[(2)] When the little girl attacked the lamb remained calm.
\end{itemize}
as meaning that the little girl attacked the lamb, while the `correct' interpretation derived through a full parse of the sentence is that the girl was attacked by the lamb \citep{christianson_thematic_2001,ferreira2001misinterpretations}.

Such observations have motivated the idea that language processing can be `good enough' or heuristic: comprehenders are willing to tolerate some level of inaccuracy in their interpretations in return for not having to expend too much computational effort  \citep{ferreira2002good,ferreira2007good}. More generally, language processing has been held to proceed through stages reflecting different levels of depth, with an initial shallow, heuristic, and local stage of processing, which may be followed by deeper, global, and more effortful processing when necessary \citep{trueswell1993verb,frazier1982making,macdonald_lexical_1994,gouvea2010linguistic,hoeks_seeing_2004,van_gompel_unrestricted_2000,bever1970cognitive,frazier1978sausage,tabor2004effects}.

While the idea of shallow processing explains a wide variety of phenomena and has intuitive appeal, the idea of `depth of processing' has yet to be made precise in a way that links it formally to theories of computation under resource constraints \citep{lewis2014computational,lieder2020resource}. We propose a computational model in which language comprehension instantiates optimal tradeoffs between accuracy and processing depth, where processing depth is measured information-theoretically in terms of bits of information extracted from perceptual input.   Our model joins recent explanations of human and animal perception and behavior based on rate--distortion theory \citep{sims2016rate,sims2018efficient,zaslavsky2018efficient,zaslavsky2020rate,mollica2021forms,cheyette2020unified,bhui2021resource,arumugam2022rate,futrell2023information}. Furthermore, our model provides a link between notions of depth of processing, processing effort, and processing time: we hold that processing depth increases with time, and that an increase in processing depth is reflected in processing effort, which in turn is reflected in EEG signals and reading times.

Below, we present our model using comprehension of Example~(1) as a running example. Then we validate the model by simulating N400 and P600 effects elicited by a variety of complex semantic and syntactic manipulations from two experiments, and by simulating reading times for garden-path constructions in the Syntactic Ambiguity Processing dataset \citep{huang2023surprisal}, successfully predicting the relative magnitudes of garden path slowdowns in reading time.

\section{Model}

\begin{figure}
\centering
\includegraphics[width=.35\textwidth]{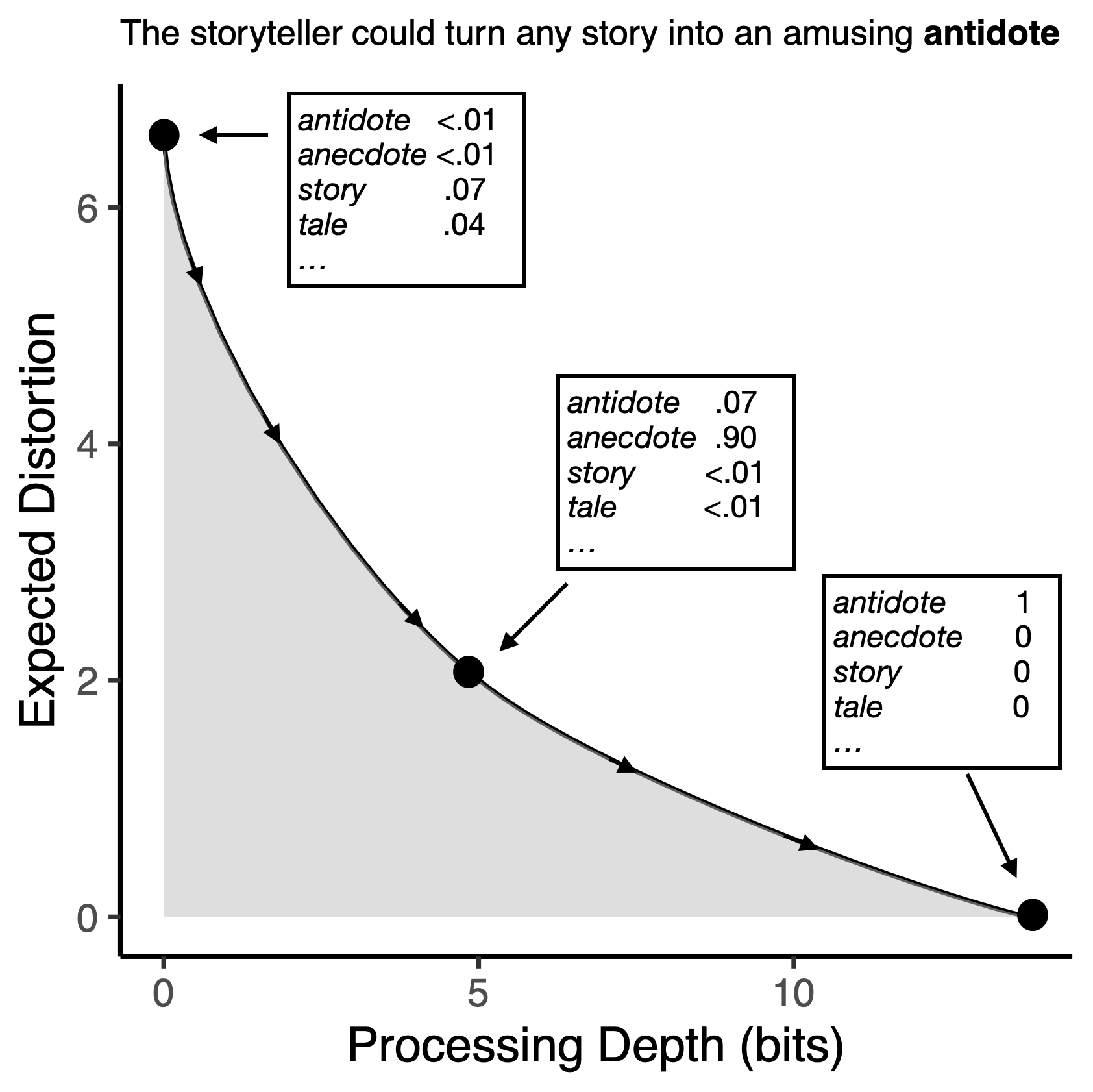}
\caption{Tradeoff between distortion and processing depth (KL divergence) in optimal interpretation policies for the given input. Each location in the white part of the plane represents a possible interpretation policy for the input; the tradeoffs in the gray region are unachievable. The black line shows the \emph{efficient frontier} of policies that achieve the minimal distortion for a given level of processing depth. We hold that interpretation policies move down this frontier with increasing processing time.}
\label{fig:rd-policy-curve}
\end{figure}

We characterize online language comprehension using an \key{interpretation policy}, which represents a comprehender's probability distribution on interpretations $w$ given perceptual inputs $x$. We hold that comprehenders' interpretation policies minimize a \textbf{distortion metric}---a measure $d(w,x)$ of how bad it is to form an interpretation $w$ given input $x$---subject to a constraint on bits of information extracted from the sensory input.  An example of interpretation policies and their tradeoffs is shown in Figure~\ref{fig:rd-policy-curve}. We furthermore posit that the tradeoff of distortion and processing depth depends on time: while a comprehender is perceiving input $x$, more information becomes available to them, and so their interpretation policy moves along the efficient frontier in Figure~\ref{fig:rd-policy-curve}, evolving from a shallow, inaccurate policy to a a deep, accurate policy while maintaining an optimal tradeoff.%,cheyette2020unified,gershman2020origin,bhui2021resource,arumugam2022rate

\paragraph{Formal model} We propose that the comprehenders' interpretation policy $p_t(w|x)$ at a given time $t$ is chosen to minimize the expected distortion subject to a constraint on the depth of processing, $D(t)$:
\begin{align}
\label{eq:opt}
&\mathop{\text{minimize}}_{p_t(w \mid x)} \mathop\mathbb{E}_{p(x)p_t(w \mid x)}\left[d(w,x)\right] \text{ subject to }D(t) \le C(t),
\end{align}
where the processing depth $D(t)$ is the average KL divergence \citep{cover2006elements} between the interpretations $p_t$ and the `default' policy $p_0(w)$ representing a comprehender's prior expectations:
\begin{equation}
\label{eq:kl}
D(t) = \mathop\mathbb{E}_{p(x)}\left[D_{\text{KL}}\left[p_t(w \mid x) \| p_0(w) \right]\right],
\end{equation}
and $C(t)$ is a constraint on the processing depth achievable at time $t$ \citep[][]{cheyette2020unified}. The KL divergence reflects the extent to which the interpretation policy $p_t(w \mid x)$ moves away from the default interpretations $p_0(w)$ in response to the information provided by the input $x$. 

To summarize, we are optimizing the processing accuracy (minimizing distortion) in Eq.~\ref{eq:opt} with a processing depth constraint as defined in Eq.~\ref{eq:kl}. Applying the method of Lagrange multipliers, the optimal interpretation policies have the form
%%reviewer 1 also suggested us to write out normalizing constant for eq2 and eq3, but I feel like it might not be necessary
\begin{equation}
    \label{eq:heuristic-policy}
    p_t(w \mid x) = \frac{1}{Z_\lambda(x)} p_0(w) e^{-\lambda(t) d(w, x)},
\end{equation}
where $Z_\lambda(x)$ is a normalizing constant. We posit that time-varying function $\lambda(t)$ has the form $\lambda(t) = \lambda_0 t$ for constant $\lambda_0$ (thus implicitly defining the time-varying depth constraint $C(t)$). This stipulation is inspired by models of perception as noisy evidence accumulation \citep{bogacz2006physics,norris2008perception,bicknell2010rational}, where a prior $p_0(w)$ is continually updated in response to noisy perceptual evidence. The final interpretation policy which we use in this paper is
\begin{equation}
\label{eq:the-policy}
p(w \mid x) \propto p_0(w) e^{-\lambda_0 t d(w, x)}.
\end{equation}

To give some intuition for model behavior: at time $t=0$, the interpretation policy reduces to $p_t(w \mid x) = p_0(w)$, meaning that a comprehender's interpretations are based on prior expectations, taking into account no information from the input. As the time $t \rightarrow \infty$, the policy $p_t(w \mid x)$ concentrates all probability mass on interpretations with minimum distortion, that is, on veridical interpretations, at which point the processing depth is equal to the surprisal of the input, $D(t) = -\log p(x)$, indicating that all the information in the input $x$ has been processed.

\begin{figure}
\includegraphics[width=.5\textwidth]{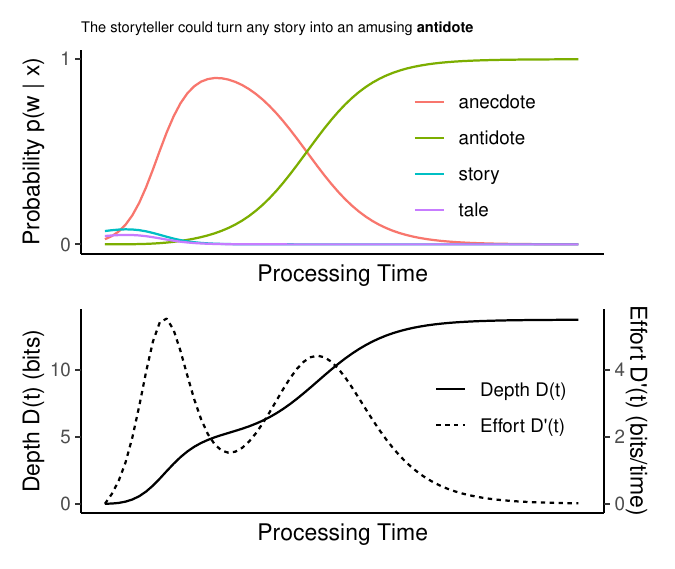} 
\caption{\textbf{Top.} Probabilities of four interpretations $w$ given input $x=\text{``story''}$ in the given context as a function of processing time, as predicted from Eq.~\ref{eq:the-policy} with parameters described in the text. \textbf{Bottom.} Measures of processing effort over time for the same input.}
\label{fig:opt-interp}
\end{figure}

\begin{figure}
\centering
\includegraphics[width=.45\textwidth]{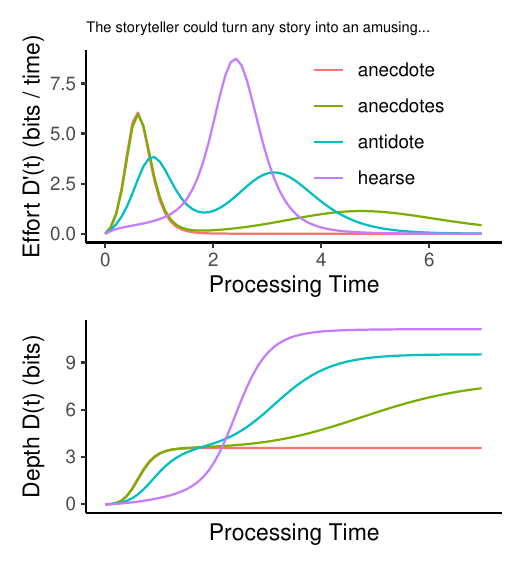} 
\caption{Processing timecourses for four different inputs in the given context. ``Anecdote'' is the control and the most likely completion. ``Hearse'' represents a semantic anomaly, ``anecdotes'' represents a syntactic anomaly, and ``antidotes'' represents a recoverable anomaly---semantically anomalous input that can be easily mistaken for a more likely input.}
\label{fig:four_alt}
\end{figure}

\paragraph{Application to language comprehension} We represent the inputs $x$ and interpretations $w$ as strings. We use a distortion measure that reflects a mixture of semantic and form-based distance between $w$ and $x$:
\begin{equation}\label{eq:likelihood}
d(w, x) = d_\varphi(w, x) + \gamma d_\sigma(w,x), 
\end{equation}
where $d_\varphi$ is a form-based phonological or orthographic distance metric, $d_\sigma$ is a semantic distance metric, and $\gamma$ is a scalar controlling the relative importance of semantic as opposed to form-based distance. We implement the form-based distance as orthographic edit distance and semantic distance as cosine distance between GPT-2 word embeddings \citep{radford2019language}. For the default policy $p_0(w)$, we use the GPT-2 language model \citep{radford2019language}.

Figure~\ref{fig:opt-interp} shows model behavior using this setup for the last word in Example~(1). For this example, we use only form-based distortion (that is, $\gamma=0$), and the set of possible interpretations $w$ ranges over the 10,000 most frequent words in SUBTLEXus \citep{brysbaert2009moving}. Initially, when no perceptual input is taken into account, ``story'' is the most likely interpretation, as it is the most likely continuation in the context under the GPT-2 language model. With short processing time, ``anecdote'' appears most likely, as it is highly expected in context and also close in distortion to the true input. With deeper processing, all probability mass is concentrated asymptotically on the veridical ``antidote''. The shift from ``anecdote'' to ``antidote'' happens when processing depth reaches around 5 bits. 

\paragraph{Processing effort} So far our model describes the evolution of a comprehender's interpretations of their input with time. We require a link between this model and measures of processing effort, such as EEG amplitude and reading time. We propose that processing effort corresponds to \emph{change} in processing depth: it requires effort to increase depth. We define the instantaneous \key{effort} at time $t$ as the rate of change of processing depth $D(t)$, given by the time derivative $D^\prime(t) = \frac{d}{dt} D(t)$. %The instantaneous effort $D^\prime(t)$ is always positive, meaning that processing can only become deeper over time.

Example timecourses for depth $D(t)$ and effort $D^\prime(t)$ are shown in Figure~\ref{fig:opt-interp}, bottom. Here, the early concentration of probability mass on ``anecdote'' is reflected in an initial pulse in the effort $D^\prime(t)$. The later shift of probability mass from ``anecdote'' onto the veridical ``antidote'' creates a second pulse. Thus, processing \emph{appears} to proceed in two distinct stages: first a `shallow' stage of processing, followed by a later `deep' error correction. This behavior is in fact generated by a single process of continuously increasing processing depth with time. 
 
\paragraph{Link to EEG measures} The idea of shallow-to-deep language processing has also been used to explain EEG signals of language comprehension, with the N400 signal a function of effort in shallow processing and the later P600 signal a signal of anomalies detected through deep processing \citep{kutas_event-related_1980,hagoort_syntactic_1993,kim_independence_2005,van_herten_erp_2005,van_herten_when_2006,ito2016predicting,kuperberg_neural_2007,kuperberg2020tale,van2012prediction,kolk2003structure,hoeks_seeing_2004}. In support of this idea, anomalous words such as those in Example~(1), which are surprising but close in form to a plausible alternative, yield a biphasic N400 and P600 effect \citep{ryskin2021erp}.

To link our measures to EEG data, we posit the voltage $V(t)$ of EEG signal at time $t$ is proportional to the  effort $D^\prime(t)$ modulated by a carrier wave with angular frequency $\omega$ and phase $\phi$, as
\begin{equation}
    \label{eq:eeg}
  V(t) \propto - D^\prime(t) \sin(\omega t + \phi). 
\end{equation}
In this view, the N400 and P600 signals are manifestations of a singular underlying process of increasing processing depth. 
N400 arises when there is a large early pulse in effort, and P600 arises when there is a late pulse. 

\begin{figure}
\centering
\includegraphics[width=.4\textwidth]{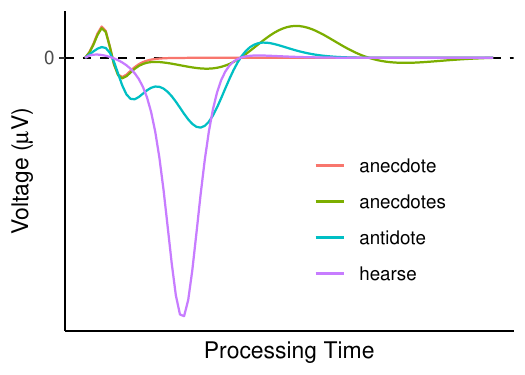} 
\caption{Simulated EEG signal corresponding to the processing timecourses in Figure~\ref{fig:four_alt}, using Eq.~\ref{eq:eeg} with $\omega=1$ and $\phi=-\nicefrac{2}{3}$.}% Displayed are the differences in voltage to the control condition (``anecdote'').}
\label{fig:eeg}
\end{figure}

In support of this idea, Figure~\ref{fig:four_alt} shows processing timecourses for four different inputs used in \cite{ryskin2021erp}: a control input ``anecdote'', a syntactically anomalous input ``anecdotes'', a semantically anomalous input ``hearse'', and a recoverable semantic anomaly ``antidote'' that is phonologically close to the control. The semantic anomaly yields a single, large, delayed pulse in processing effort. The delay happens because early processing for this input fails to concentrate probability mass on any particular interpretation---``hearse'' is initially so unlikely under $p_0(w)$ that a great deal of evidence has to be accumulated for its probability to rise. In contrast, the syntactic and recoverable semantic anomalies create two pulses of processing effort. For the syntactic anomaly, the timecourse of processing is nearly identical to the control input up to a depth of around 5 bits, at which point there is a very late and temporally diffuse second pulse of processing effort, corresponding to the P600 signal. Mapping the processing effort onto predicted voltages following Eq.~\ref{eq:eeg} in Figure~\ref{fig:eeg}, we predict (1) early negativity for the semantic and recoverable semantic anomalies, and (2) late positivity for the syntactic and recoverable semantic anomalies. This pattern matches the empirical findings.

\paragraph{Link to reading times} To capture reading times, we posit that comprehenders move to the next word when they judge that deeper processing is no longer necessary. Concretely, we model reading times as the minimum amount of time needed for processing effort to fall below a threshold $\varepsilon$. It is thus possible that a reader may move on to the next word before fully resolving the current word, as long as the \emph{change} in processing depth with time is small enough. For example, in the processing timecourse for the syntactically anomalous ``anecdotes'' in Figure~\ref{fig:four_alt}, a reader may move on during the trough between the two pulses in processing effort, if the processing effort falls below the threshold $\varepsilon$ at this point. In that case the reader would have a lingering misinterpretation.

\paragraph{Relation to noisy-channel models} Our shallow interpretation policy generalizes noisy channel models of language comprehension \citep{gibson_rational_2013,ryskin2021erp,poppels2016structure}. The policy reproduces the predictions of noisy-channel models in the special case when $\lambda(t) = 1$ and the distortion metric is equal to a log-likelihood under a noise model, i.e. $d(w, x) = -\ln p_N(x \mid w)$ for some noise model $p_N$. It differs from noisy channel models in its cognitive interpretation and in its generality: in the shallow processing model, the distortion $d(w,x)$ need not be a log-likelihood (that is, it need correspond to a normalized probability $p_N(x \mid w)$). Increasing $t$ in Eq.~\ref{eq:the-policy} corresponds to accumulation of samples of evidence in a noisy-channel model of perception \citep{norris2008perception,bicknell2010rational}.

\section{Study 1: N400, P600, and biphasic EEG signals}
\begin{table}[!bhtp]
\centering
\resizebox{\linewidth}{!}{%
\begin{tabular}{lllll}\toprule
Exp & Condition & Context & Target &ERPs  \\\midrule
\multirow{4}{*}{Ito-16} & SemRelated & \multirow{4}{*}{\parbox{2.1cm}{The student is going to the library to borrow a \dots}} & page & Reduced N400 \\
 & FormRelated & & hook & Biphasic \\
 & Unrelated & & sofa & N400 \\
 & Control & & book & NA \\\midrule
 \multirow{4}{*}{{Ryskin-21}} & Semantic & \multirow{4}{*}{\parbox{2.1cm}{The storyteller could turn any incident into an amusing \dots}}& hearse & N400 \\
 & Syntactic & & anecdotes & P600 \\
 & SemCrit & & antidote & Biphasic \\
 & Control & & anecdote & NA \\\bottomrule
\end{tabular}}%
\caption{List of conditions, sample sentences and ERP patterns in dataset.}
\label{tab:eeg_stimuli}
\end{table}

\paragraph{Dataset}
We validate our theory on two EEG experiments, featuring N400 effect, P600 effect and biphasic effect. Table~\ref{tab:eeg_stimuli} shows a list of conditions with sample stimuli and empirical ERP patterns across experiments. The ERP effects in the experimental conditions are all calculated in terms of differences to the EEG amplitude in the control condition. The experiment \emph{Ito-16} includes three different kinds of semantic violations \citep{ito2016predicting} that are semantically related (\emph{SemRelated}), orthographically related (\emph{FormRelated}) or unrelated (\emph{Unrelated}) to the predicable target. The N400 effects in \emph{FormRelated} and \emph{SemRelated} conditions are smaller than in the \emph{Unrelated} conditions, and only \emph{FormRelated} condition elicits a P600 effect. The experiment \emph{Ryskin-21} introduces semantic violations (\emph{Semantic} condition), syntactic violations (\emph{Syntactic} condition) and semantic violations that are phonologically similar to the predictable control target (\emph{SemCrit} condition). There is a biphasic N400--P600 effect in the \emph{SemCrit} condition, but the effect size is smaller than the N400 in the \emph{Semantic} condition and P600 in the \emph{Syntactic} condition.

\paragraph{Implementation}
We calculate interpretation policies over time for the critical word in each sentence, with the set of alternative interpretations $w$ consisting of the words in all conditions. We set the parameter $\gamma$ by visual inspection, obtaining $\gamma=10$ for Ryskin-21 and $\gamma=6$ for Ito-16.

\begin{figure}[!htb]
    \centering
    \begin{subfigure}{.5\textwidth} % Adjust the width as needed
        \includegraphics[width=\linewidth]{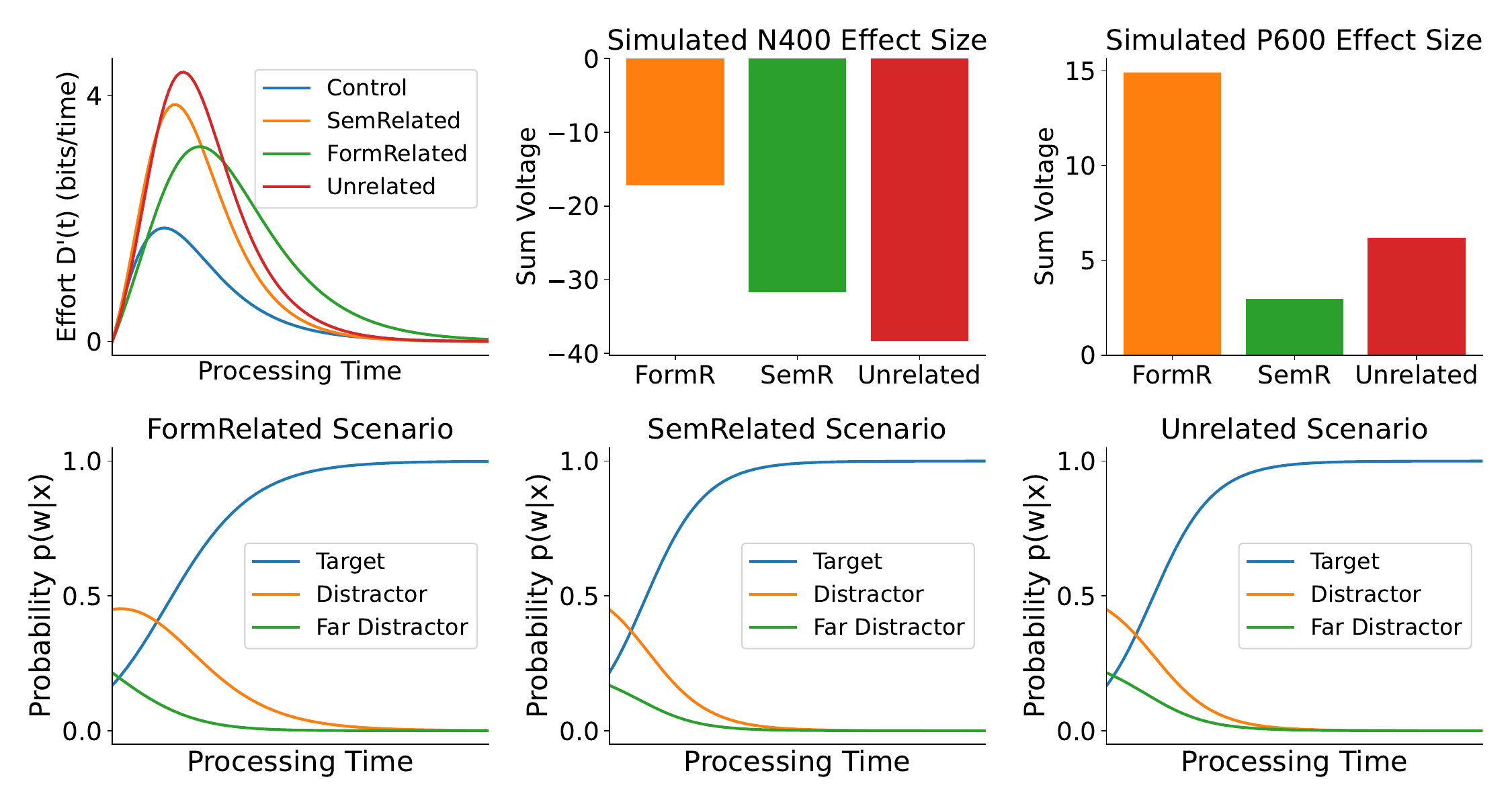} % Replace with your image file
        \caption{Ito-16 experiment}
        \label{fig:ito}
    \end{subfigure}
    \hfill
    \begin{subfigure}{.5\textwidth} % Adjust the width as needed
        \includegraphics[width=\linewidth]{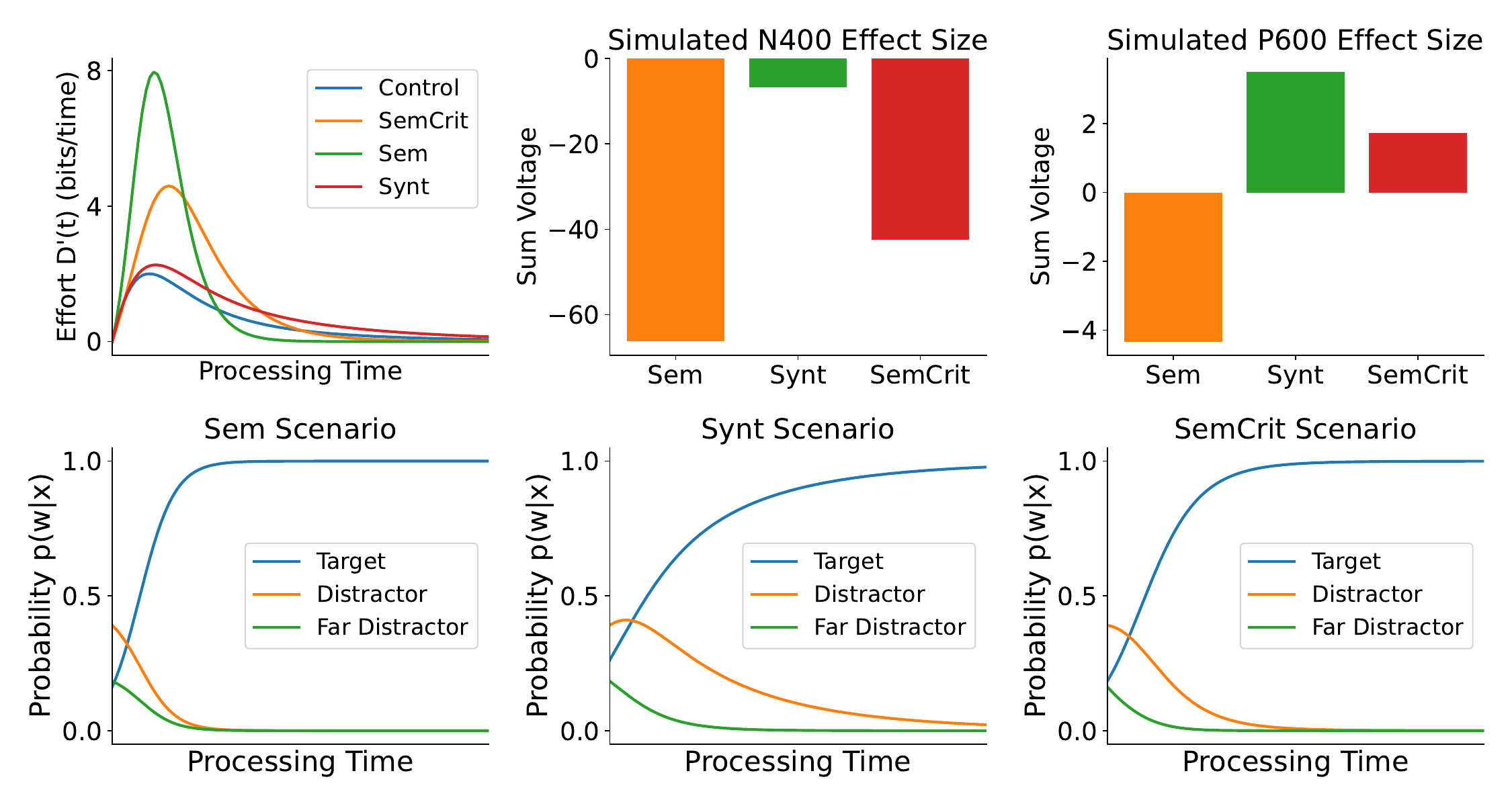} % Replace with your image file
        \caption{Ryskin-21 experiment}
        \label{fig:ryskin}
    \end{subfigure}
    \caption{Simulation results for EEG experiments. \textbf{Left.} Instantaneous processing effort ($D'(t)$) \textbf{Middle.} Simulated N400 effect size. \textbf{Right.} Simulated P600 effect size.}
    \label{fig:eeg_result}
\end{figure}

\paragraph{Results}
Figure~\ref{fig:eeg_result} shows that our model simulates the observed EEG patterns in both Ito-16 and Ryskin-21. In Ito-16, our model predicts a graded N400 effects elicited by three kinds of semantic violations (\emph{Unrelated} $>$ \emph{SemRelated} $>$ \emph{FormRelated}), where semantic violations that are either orthographically nor semantically related to the predicted word trigger the largest N400 responses. Our model also predicts a greater P600 amplitude in \emph{FormRelated} than other conditions. In Ryskin-21, our model predicts the order the effect size across conditions: a graded N400 effect (\emph{Semantic} $>$ \emph{SemCrit} $>$ \emph{Syntactic}) and a graded P600 effect (\emph{Syntactic} $>$ \emph{SemCrit} $>$ \emph{Semantic}). 

The model behavior can be explained in terms of the different rates of probabilistic distribution shift: the semantically anomalous conditions with a large N400 effect tend to have the fastest probability update, and anomalous conditions with a large P600 effect (e.g. \emph{Syntactic} condition in Ryskin-21; \emph{FormRelated} condition in Ito-16) has the lowest probability update rate. The different probability shift rate results in the increase of instantaneous effort to appear early or delayed.

\section{Study 2: Garden path reading times}

\begin{table}[bth]
    \centering
    \resizebox{\linewidth}{!}{%
    \begin{tabular}{p{0.15\linewidth} p{0.85\linewidth}}
    \toprule Type & Sentence\\\midrule
    \multirow{2}{*}{MV/RR} & The little girl fed the lamb \textbf{\underline{remained}} relatively calm\dots \\
     & The little girl who was fed the lamb \textbf{\textit{remained}} relatively calm\dots \\\midrule
    \multirow{2}{*}{NP/S}  & The little girl found the lamb \textbf{\underline{remained}} relatively calm. \\
    & The little girl found that the lamb \textbf{\textit{remained}} relatively calm\dots\\\midrule
    \multirow{2}{*}{NP/Z} & When the little girl attacked the lamb \textbf{\underline{remained}} relatively calm\dots\\
    & When the little girl attacked, the lamb \textbf{\textit{remained}} relatively calm\dots\\\bottomrule
    \end{tabular}
    }
    \caption{A list of garden path constructions and example sentences. Critical words are marked in \textbf{bold}. The critical words in garden path sentences are \underline{underlined}, and the control counterparts are \textit{italic}.}
    \label{tab:garden-path-stimuli}
\end{table}

\begin{figure}[!thb]
    \centering
    \includegraphics[width=\linewidth]{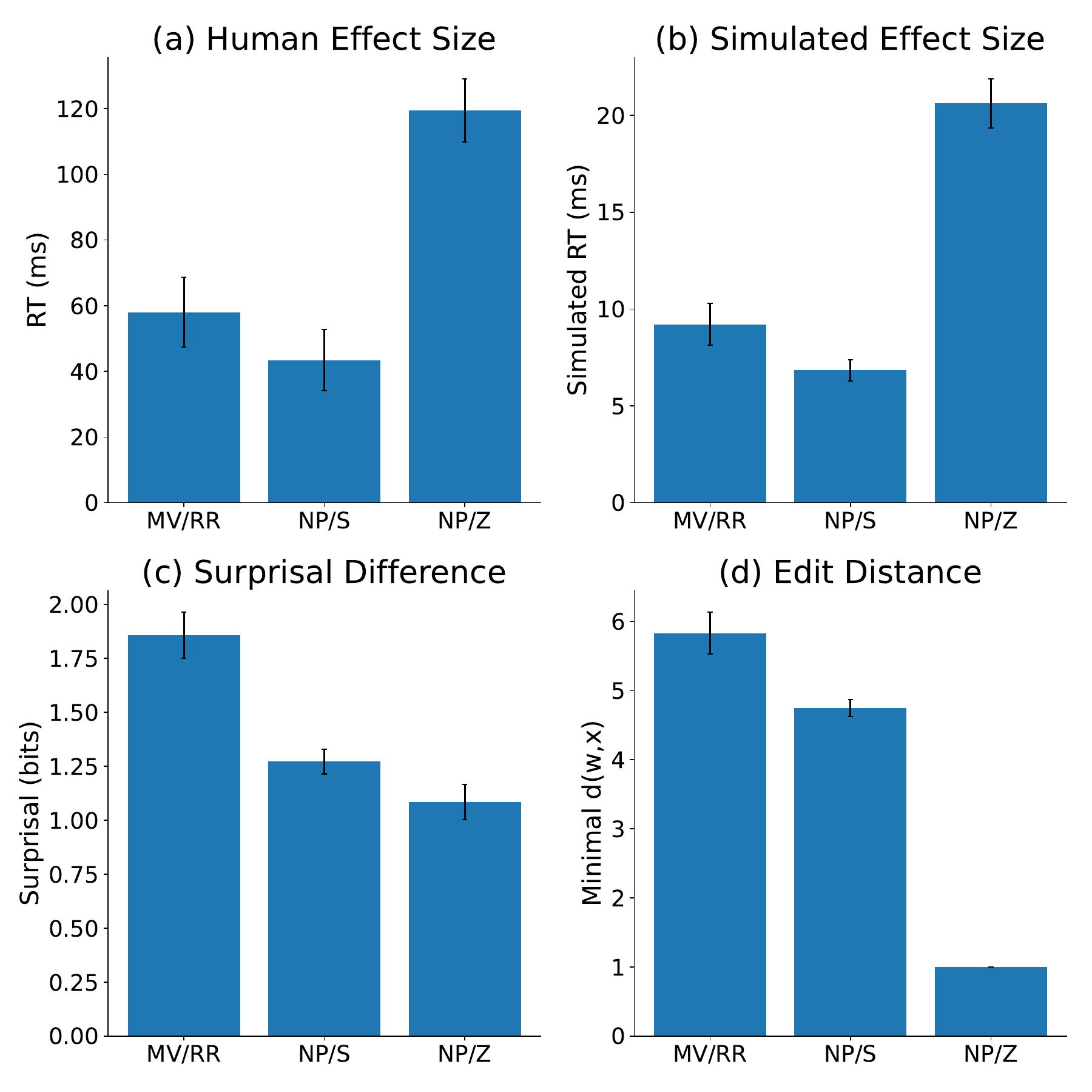}
    \caption{
    \textbf{(a)} Average effect size (RT difference to unambiguous control) at critical region estimated from human experiment.  % RF: For the target word or spillover? % For target word
    \textbf{(b)} Average effect size estimated from model simulation. \textbf{(c)} Average surprisal difference between experimental and control sentences. 
    \textbf{(d)} Average edit distance between presented sentence and closest alternative. Error bars show standard errors.}
    \label{fig:sap_bar}
\end{figure}

\paragraph{Dataset} We used a set of three classic garden path constructions from large-scale syntactic ambiguity processing benchmark, consisting of self-paced reading time data from 2000 participants~\citep{huang2023surprisal}. The constructions include: Main verb/ reduced relative clause garden path (MV/RR), Direct object/ sentential complement garden path (NP/S), Transitive/ intransitive garden path (NP/Z) (see Table~\ref{tab:garden-path-stimuli}). There are 24 lexically matched item sets for each construction. Each set has an experimental sentence with a tempory syntactic ambiguity and an unambiguous control sentence. %We define the garden-path effect as the reading time difference at the critical disambiguation region between the temporarily ambiguous experimental sentence and the unambiguous control counterpart. 
These constructions generate reliable garden path effects \citep{frazier1979comprehending,bever1970cognitive,grodner2003against,sturt1999structural}, but differ in the magnitude of the effect \citep{van2021single,huang2023surprisal}. In particular, NP/Z constructions have the largest effect size at the critical region, followed by MV/RR and NP/S: see Figure~\ref{fig:sap_bar}(a). %The order of effect magnitude cannot be explained by the surprisal difference between experimental and control conditions, where the MV/RR has the largest surprisal difference followed by NP/S and NP/Z constructions (see Fig. \ref{fig:sap_bar} (c)). 

\paragraph{Implementation}
We consider interpretations $w$ ranging over the entire sentence, rather than only the critical word, with a set of alternatives consisting of all sentences in a lexically matched set across different constructions as possible interpretations given a single sentence.  This simulates the scenario where a comprehender is re-evaluating and updating their beliefs about the entire sentence in parallel \citep{wen2021transposed,wen2021fast}. The prior $p(w)$ is calculated as the sum of conditional log probability for every word in the sentence from GPT-2. We set $\gamma=0$ because all alternatives are lexically matched, and set the reading time threshold to $\varepsilon=.001$.
%Given that all candidate interpretations are lexically matched, we set $\gamma = 0$. %We compared human reaction time effect size at critical region with simulated effect size. We define the effect size as the difference between experimental and control conditions in each construction, and we compared our simulation results with simulation based surprisal difference.

\begin{figure}[!htb]
    \centering
    \includegraphics[width = \linewidth]{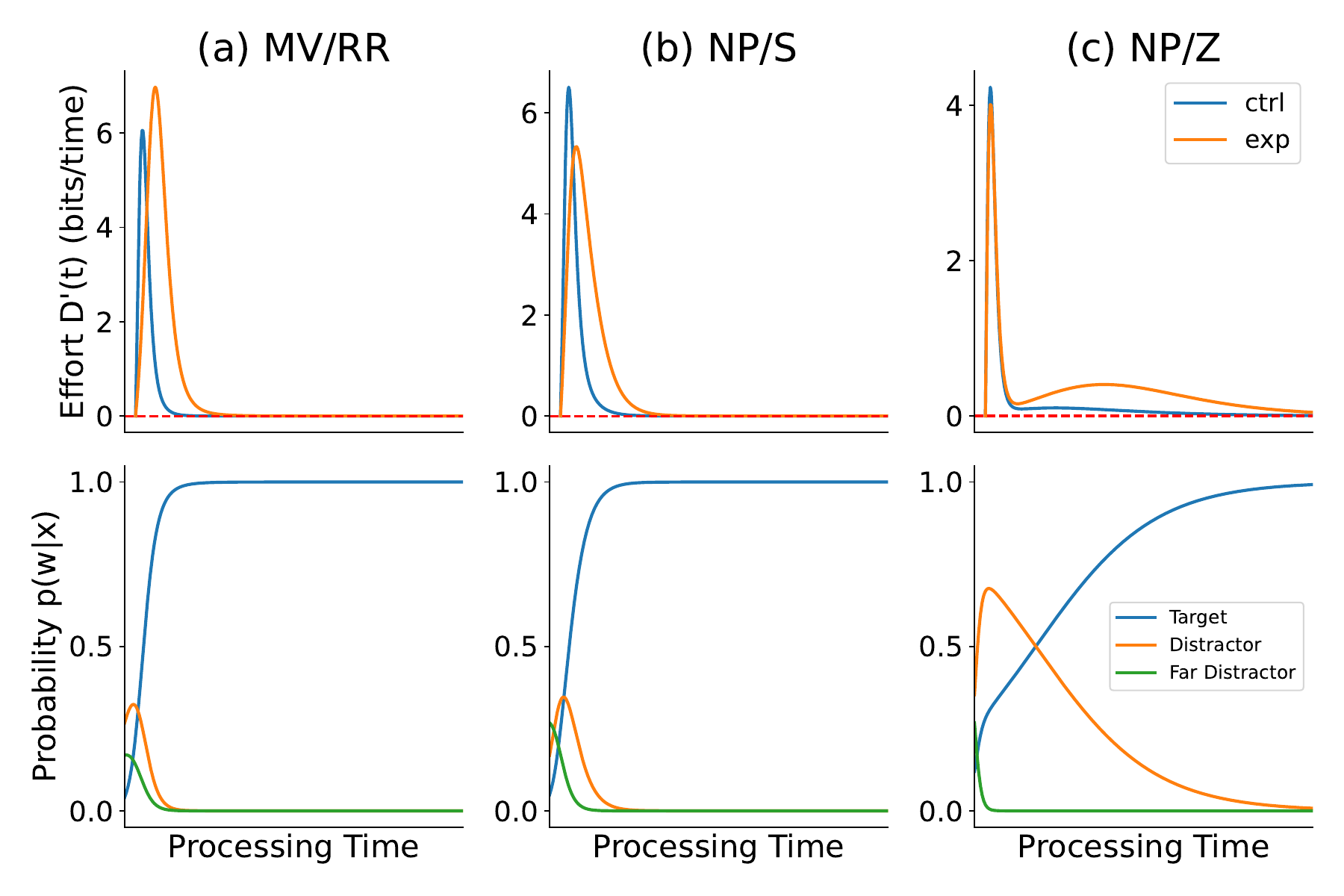}
    \caption{\textbf{Top.} Instantaneous processing effort over time for (a) MV/RR; (b) NP/S and (c) NP/Z constructions. The dotted red line indicates the reading time threshold $\varepsilon = .001$. \textbf{Bottom.} Probabilities of three different interpretations $w$ given an experimental sentence (\emph{Target}). \emph{Distractor} is the counterpart control sentence and \emph{Far Distractor} is a sentence with a different syntactic construction.}
    \label{fig:sap_timecourse}
\end{figure}

\paragraph{Results}
Figure~\ref{fig:sap_bar} shows that our simulation replicates the relative magnitude of the three garden path effects. This result cannot be explained by the surprisal difference between experimental and control conditions, where the MV/RR has the largest surprisal difference followed by NP/S and NP/Z, nor by the distortion (edit distance) alone: see Figure~\ref{fig:sap_bar}(c--d).

The model captures the reading time differences among garden paths because the reading time is affected by the comparative strength of different candidate interpretations, which is jointly decided by both prior expectations and similarity to the input. In Figure~\ref{fig:sap_timecourse}(a--b), we see that for MV/RR and NP/S, the veridical interpretation initially has a very low probability, but the probability increases rapidly because the distance to the more-likely control interpretation is large. On the other hand, for NP/Z---shown in Figure~\ref{fig:sap_timecourse}(c)---the veridical interpretation is not extremely low probability initially, but the probability for this interpretation grows very slowly because there is only weak evidence for it: the distance between the veridical input and the closest distractor (the unambiguous variant with a comma) is very small.

\section{Discussion}
We have proposed a unified computational model of language comprehension as a function of depth of processing. In the model, shallow interpretations are generated based on a trade-off between accuracy and the amount of information extracted from perceptual input. We quantify reading time and EEG signals as two different indices of changes in comprehenders' beliefs. Our model provides a simple and transparent explanation for the order of effect size across three different types of garden path constructions, and successfully simulates certain ERP effects.

%Our model bridges behavioral and neural signatures of language comprehension. We hypothesize that EEG signals could be mapped to instantaneous processing effort modulated by a neural oscillation, whereas reading time reflects the amount of time for comprehenders to achieve a stable probabilistic representation of their input. The formal link between reading time and EEG signals calls for more co-registration studies in the field.

Our model quantifies the well-known hypothesis of a heuristic ``shallow processing'' mechanism in a time-sensitive and dynamic way. Rather than positing two discrete stages of shallow and then deep processing, or that processing involves separate streams of heuristic and veridical input \citep{kim_independence_2005,van_herten_erp_2005,van2011monitoring}, our model posits a single interpretation process with continuously increasing in processing depth. For certain inputs, this process creates dynamics with two pulses of processing effort, corresponding to formation of a shallow interpretation followed by error detection or correction. %Thus, our model expands the predictive power of single-stream/one-stage models to account for processing of inputs that require additional time or effort to resolve. 

Our model enhances theories of processing difficulty based on expectations by introducing a new factor in the form of distortion. Our model's indices of processing effort are closely related to the surprisal of the input, which has been proposed as a predictor of reading times \citep{hale2001probabilistic,levy2008expectation} and EEG amplitudes \citep{frank2015erp,michaelov2024strong}. The veridical interpretation's initial probability at time $t=0$ is determined entirely by prior expectations, and the total cumulative effort to process a word is equal to its surprisal. Our model differs from Surprisal Theory in that reading time is also influenced by distances to distractor interpretations. These distances determine the \emph{rate} at which the probability of the correct interpretation increases, yielding potentially dramatic slowdowns when that rate is small. Our model is also compatible with theories of ERPs where the N400 reflects the surprisal of a likely but non-veridical interpretation, P600 reflects reconciliation of interpretations, and the total cumulative amplitude reflects surprisal \citep{li2023heuristic,li2023decomposition}.

While we have demonstrated initial success for our model, it has some limitations. Most serious is the representation of possible interpretations $w$ as strings, with the concomitant use of edit distance (augmented with semantic distance) for the distortion. In a more complete model implementation, it is possible that interpretations $w$ may range over structured objects such as syntactic treelets, with distortion reflecting a structure-sensitive distance function among these objects. Future work will determine the proper structure and distortion measure for comprehenders' interpretations.

Our model and results provide evidence that processing constraints in language comprehension can be captured information-theoretically in terms of rational use of computational resources \citep{hahn2022resource}, joining with similar models in the domains of language production \citep{futrell2023information}, semantics \citep{zaslavsky2018efficient,steinert2020quantifiers,imel2022modal}, and pragmatics \citep{zaslavsky2020rate,zhou2022teasing}, as well as other areas of cognition \citep{bhui2021resource,arumugam2022rate}.

\printbibliography 
\end{document}